\documentclass[a4paper,conference]{IEEEtran}
\IEEEoverridecommandlockouts
\usepackage{cite}
\usepackage{amsmath,amssymb,amsfonts}
\usepackage{algorithmic}
\usepackage{graphicx}
\usepackage{textcomp}
\usepackage{xcolor}
\def\BibTeX{{\rm B\kern-.05em{\sc i\kern-.025em b}\kern-.08em
    T\kern-.1667em\lower.7ex\hbox{E}\kern-.125emX}}

\graphicspath{{figures/}}
\DeclareGraphicsExtensions{.pdf,.jpeg,.png}

\usepackage{url}
\usepackage{xspace}
\usepackage{color}
\usepackage{cleveref}
\usepackage{booktabs}
\usepackage{multirow}
\usepackage{subfigure}
\usepackage{balance}
\definecolor{orange}{rgb}{1,0.3,0}

\newif\ifinternalreview

\internalreviewfalse

\newcommand{\methodname}{PolyLaneNet}

\ifinternalreview

    \newcommand{\NOTE}[1]{\textcolor{blue}{NOTE: #1}}
    \newcommand{\Berriel}[1]{\textcolor{magenta}{Berriel: #1\xspace}}
    \newcommand{\px}[1]{\textcolor{red}{PX: #1\xspace}}
    \newcommand{\thiago}[1]{\textcolor{green}{Thiago: #1\xspace}}
    \newcommand{\pxst}[1]{\textcolor{red}{PX: \st{#1}\xspace}}
    \newcommand{\Tabelini}[1]{\textcolor{orange}{Tabelini: #1\xspace}}
\else

    \newcommand{\NOTE}[1]{}
    \newcommand{\Berriel}[1]{}
    \newcommand{\px}[1]{}
    \newcommand{\thiago}[1]{}
    \newcommand{\pxst}[1]{}
    \newcommand{\Tabelini}[1]{}
\fi

\begin{document}

\title{\methodname: Lane Estimation\\via Deep Polynomial Regression}

\author{
    \IEEEauthorblockN{
        Lucas Tabelini\IEEEauthorrefmark{1},
        Rodrigo Berriel\IEEEauthorrefmark{1},
        Thiago M. Paixão\IEEEauthorrefmark{1}\IEEEauthorrefmark{2},
        Claudine Badue\IEEEauthorrefmark{1},\\
        Alberto F. De Souza\IEEEauthorrefmark{1} and
        Thiago Oliveira-Santos\IEEEauthorrefmark{1}
    }
    \IEEEauthorblockA{
        \IEEEauthorrefmark{1}Universidade Federal do Espírito Santo (UFES), Brazil\\
        \IEEEauthorrefmark{2}Instituto Federal do Espírito Santo (IFES), Brazil\\
        {\small \texttt{Email: tabelini@lcad.inf.ufes.br}}
    }
}

\maketitle
\begin{abstract}
	One of the main factors that contributed to the large advances in autonomous driving is the advent of deep learning. For safer self-driving vehicles, one of the problems that has yet to be solved completely is lane detection. Since methods for this task have to work in real-time (+30 FPS), they not only have to be effective (i.e., have high accuracy) but they also have to be efficient (i.e., fast). In this work, we present a novel method for lane detection that uses as input an image from a forward-looking camera mounted in the vehicle and outputs polynomials representing each lane marking in the image, via deep polynomial regression. The proposed method is shown to be competitive with existing state-of-the-art methods in the TuSimple dataset while maintaining its efficiency (115 FPS). Additionally, extensive qualitative results on two additional public datasets are presented, alongside with limitations in the evaluation metrics used by recent works for lane detection. Finally, we provide source code and trained models that allow others to replicate all the results shown in this paper, which is surprisingly rare in state-of-the-art lane detection methods. The full source code and  pretrained models are available at~\url{https://github.com/lucastabelini/PolyLaneNet}.
\end{abstract}

\IEEEpeerreviewmaketitle
\section{Introduction}

Autonomous driving~\cite{badue2019arxiv} is a challenging field of research that has received a lot of attention in recent years. The perceptual problems related to this field have been immensely impacted by the advances in deep learning~\cite{possatti2019traffic,yang2020part,feng2020deep}. In particular, autonomous vehicles should be capable of estimating traffic lanes because, besides working as a spatial limit, each lane provides specific visual cues ruling the travel. In this context, the two most important traffic lines (i.e., lane markings) are those defining the lane of the vehicle, i.e., the ego-lane. These lines set the limits for the driver's actions and their types define whether or not maneuvers (e.g., lane changes) are allowed. Also, it might be useful to detect the adjacent lanes so that the systems' decisions might be based on a better understanding of the traffic scene.

Lane estimation (or detection) may seem trivial at first, but it can be very challenging. Although fairly standardized, lane markings vary in shape and colour. Estimating a lane when dashed or partially occluded lane markers are presented requires a semantic understanding of the scene. Moreover, the environment itself is inherently diverse: there may be a lot of traffic, people passing by, or it could be just a free highway. In addition, these environments are subject to several weather (e.g., rain, snow, sunny, etc.) and illumination (e.g., day, night, dawn, tunnels, etc.) conditions, which might just change while driving.

The traditional approach for the lane estimation (or detection) task consists in the extraction of hand-crafted features \cite{survey2006tits,berriel2017imavis} followed by a curve-fitting process. Although this approach tends to work well under normal and limited circumstances, it is not usually as robust as needed in adverse conditions (as the aforementioned ones). Therefore, following the trend in many computer vision problems, deep learning has recently started to be used to learn robust features and improve the lane marking estimation process \cite{scnn2018aaai,linecnn2019tits,enetsad2019iccv}. Once the lane markings are estimated, further processing can be performed to determine the actual lanes. Still, there are limitations to be tackled. First, many of these deep learning-based models tackle the lane marking estimation as a two-step process: feature extraction and curve fitting. Most works extract features via segmentation-based models, which usually are inefficient and have trouble running in real-time, as required for autonomous driving. Additionally, the segmentation step alone is not enough for providing a lane marking estimate since the segmentation maps have to be post-processed in order to output traffic lines. Further, these two-step processes might ignore global information \cite{linecnn2019tits}, which are specially important when there are missing visual cues (as in strong shadows and occlusions). Second, some of these works are performed by private companies that often (i) do not provide means to replicate their results and (ii) develop their methods on private datasets, which hinders research progress. Lastly, there is room for improvement in the evaluation protocol. The methods are usually tested on datasets from the USA only (roads in developing countries are usually not as well maintained) and the evaluation metrics are too permissive (they allow error in such a way that it hinders proper comparisons),  as discussed in Section~\ref{sec:experiments}.

In this context, methods focusing on removing the need for a two-step process further reducing the processing cost could benefit advanced driver assistance systems (ADAS) that often rely on low-energy and embedding hardware. In addition, a method that has been tested on roads other than the USA's is also of benefit to the broader community. Moreover, less permissive metrics would allow to better differentiate methods and provide a clearer overview of the methods and their usefulness.

\begin{figure*}[t]
	\centering
	\includegraphics[width=\linewidth]{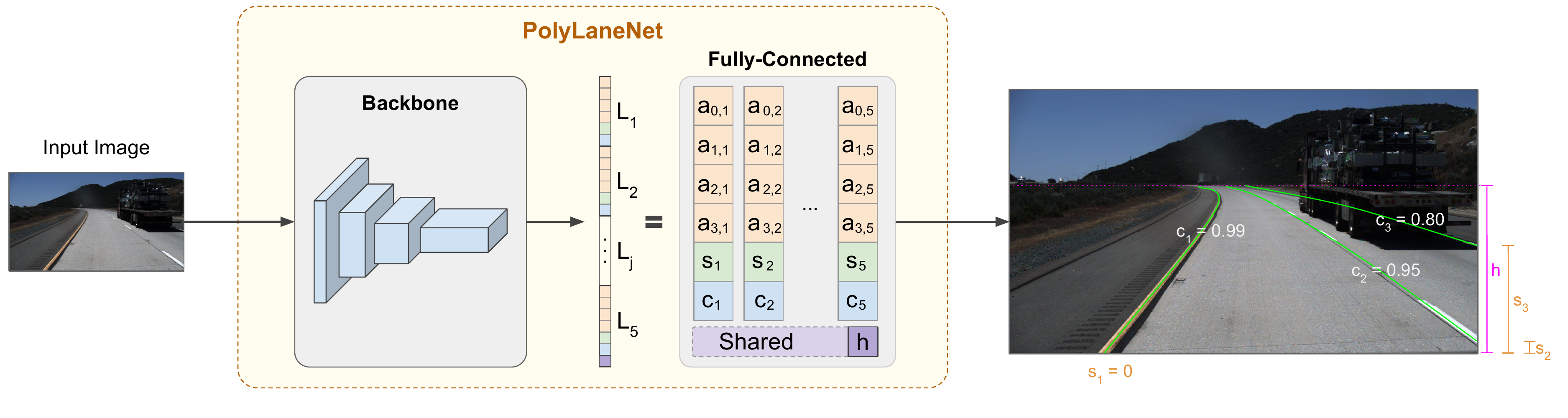}
	\caption{Overview of the proposal method. From left to right: the model receives as input an image from a forward-looking camera and outputs information about each lane marking in the image.}
	\label{fig:overview}
\end{figure*}

This work proposes \methodname{}, a convolutional neural network (CNN) for end-to-end lane markings estimation. \methodname{} takes as input images from a forward-looking camera mounted in the vehicle and outputs polynomials that represent each lane marking in the image, along with the domains for these polynomials and confidence scores for each lane. This approach is shown to be competitive with existing state-of-the-art methods while being faster and not requiring post-processing to have the lane estimates. In addition, we provide a deeper analysis using metrics suggested by the literature. Finally, we publicly released the source-code (for both training and inference) and the trained models, allowing the replication of all the results presented in this paper.

\section{Related Works}

\noindent
\textbf{Lane Detection.} Before the rise of deep learning, methods on lane detection were mostly model- or learning-based, i.e., they used to exploit hand-crafted and specialized features. Shape and color were the most commonly used features \cite{kluge1995iv,chiu2005iv}, and lanes were normally represented both by straight and curved lines \cite{jung2005imavis,berriel2015sibgrapi}. These methods, however, were not robust to sudden illumination changes, weather conditions, differences in appearance between cameras, and many other things that can be found in driving scenes. The interested reader is referred to \cite{survey2006tits} for a more complete survey on earlier lane detection methods.

With the success of deep learning, researchers have also investigated its use to tackle lane detection. Huval et al.~\cite{huval2015empirical} were one of the first to use deep learning in lane detection. Their model is based on the OverFeat and produces as output a sort of segmentation map that is later post-processed using DBSCAN clustering. They collected a private dataset on San Francisco (USA) that was used to train and evaluate their system. Because of the success of their application, companies were also interested in investigating this problem. Later, Ford released DeepLanes \cite{deeplanes2016ford}, which unlike most of the literature, detects lanes based on laterally-mounted cameras. Despite the good results, the way they modeled the problem made it less widely applicable, and they also used a private US-based dataset.

More recently, a lane detection challenge was held on CVPR'17 in which the TuSimple~\cite{tusimple-benchmark} dataset was released. The winner of the challenge was SCNN \cite{scnn2018aaai}, a method proposed for traffic scene understanding that exploits the propagation of spatial information via specially designed CNN structure. Their model outputs a probability map for the lanes that are post-processed in order to provide the lane estimates. To evaluate their system, they used an evaluation metric that is based on the IoU between the prediction and the ground-truth. After that, in \cite{linecnn2019tits}, the authors proposed Line-CNN, a model in which the key component is the line proposal unit (LPU) adapted from the region proposal network (RPN) of Faster R-CNN. They also submitted their results to the TuSimple benchmark (after the challenge was finished) with marginally better results compared to SCNN. Their main experiments, though, were with a much larger dataset that was not publicly released. In addition to this private dataset, the source code is proprietary and the authors will not release it. Another approach is FastDraw \cite{fastdraw2019cvpr} in which the common post-processing of segmentation-based methods is substituted by ``drawing'' the lanes according to the likelihood of polylines that are maximized at training time. In addition to evaluating on the TuSimple and CULane~\cite{scnn2018aaai} datasets, the authors provide qualitative results on yet another private US-based dataset. Moreover, they did not release their implementation, which hinders further comparisons. Some of the segmentation-based methods focus on improving the inference speed, as in \cite{enetsad2019iccv} (ENet-SAD) which focuses on learning lightweight CNNs by exploiting self-attention distillation. The authors evaluated their method on three well-known datasets. Although the source code was publicly released, some of the results are not reproducible\footnote{According to the author of \cite{enetsad2019iccv}, the difference in performance comes from engineering tricks neither described in the paper nor included in the available code: \url{https://web.archive.org/web/20200503114942/https://github.com/cardwing/Codes-for-Lane-Detection/issues/208}}. Closer to our work, \cite{wvangansbeke_2019} proposes a differentiable least-squares fitting module to fit a curve on points predicted by a deep neural network. In our work, we bypass the need for this module by directly predicting the polynomial coefficients, which simplifies the method while also making it faster.

In summary, one of the main problems with existing state-of-the methods is reproducibility, since most either do not publish the datasets used or the source code. In this work, we present results that are competitive with state-of-the-art methods on public datasets and fully reproducible, since we provide the source code and use only publicly available datasets (including one from outside the US).

\section{\methodname}

\noindent\\
\textbf{Model Definition.} \methodname{} expects as input images taken from a forward-looking vehicle camera, and outputs, for each image, $M_{max}$ lane marking candidates (represented as polynomials), as well as the vertical position $h$ of the horizon line, which helps to define the upper limit of the lane markings. The architecture of \methodname{} consists of a backbone network (for feature extraction) appended with a fully connected layer with $M_{max} + 1$ outputs, being the outputs $1, \ldots, M_{max}$ for lane marking prediction and the output $M_{max} + 1$ for $h$. \methodname{} adopts a polynomial representation for the lane markings instead of a set of points. Therefore, 
for each output $j$, $j=1,\ldots, M_{max}$, the model estimates the coefficients $\mathcal{P}_j=\{a_{k,j}\}_{k=0}^{K}$ representing the polynomial

\begin{equation}
    p_j(y) = \sum_{k=0}^{K} a_{k,j}y^k,
    \label{eq:pol}
\end{equation}
where $K$ is a parameter that defines the order of the polynomial. As illustrated in Figure \ref{fig:overview}, the polynomials have restricted domain: the height of the image. Besides the coefficients, the model estimates, for each lane marking $j$, the vertical offset $s_j$, and the prediction confidence score $c_j \in [0,1]$. In summary, the \methodname{} model can be expressed as

\begin{equation}
    f(I;\theta) = (\{\mathcal{P}_j, s_j, c_j)\}_{j=1}^{M_{max}},h),
\end{equation}
where $I$ is the input image and $\theta$ is the model parameters. At inference time, as illustrated in Figure \ref{fig:overview}, only the lane marking candidates whose confidence score is greater or equal than a threshold are considered as detected.

\noindent\\
\textbf{Model Training.}
For an input image, let $M$ be the number of annotated lane markings given an input image. In general, traffic scenes contain few lanes, being $M \leq 4$ for most images in the available datasets. For training (and metric evaluation), each annotated lane marking $j$, $j=1, \ldots, M$, is associated to the neuron unit $j$ of the output. Therefore, predictions related to the outputs $M + 1, \ldots, M_{max}$ should be disregarded in the loss function. An annotated lane marking $j$ is represented by a set of points $\mathcal{L}_j^*=\left\{(x^*_{i,j}, y^*_{i,j})\right\}_{i=1}^N$, where $y_{i+1,j}^* > y_{i,j}^*$, for every $i=1, \ldots,N-1$. As a rule of thumb, the higher is $N$, the more it allows to capture richer structures. We assume that the lane markings $\{\mathcal{L}_j^*\}_{j=1}^M$ are ordered according to the $x$-coordinate of the point closest to the bottom of the image, i.e., $x^*_{0, j} < x^*_{0, j+1}$ for every $j=1, \ldots, M-1$. For each lane marking $j$, the vertical offset $s^*_j$ was set as $\operatorname{min}\left(\{y^*_{i,j}\}_{i=1}^N\right)$; the confidence score is defined as

\begin{equation}
    c^*_j =
    \begin{cases}
        1, & \text{if } j \leq M\\
        0, & \text{otherwise.}
    \end{cases}
\end{equation}

The model is trained using the multi-task loss function defined as (for a single image)
\begin{equation}
\begin{aligned}
	L(\{\mathcal{P}_j\},h,\{s_j\},\{c_j\}) = & W_p L_p(\{\mathcal{P}_j\}, \{\mathcal{L}_j^*\}) \\ + & W_s \frac{1}{M}\sum_{j}L_{reg}(s_j,s_j^*) \\
	+ & W_c \frac{1}{M}\sum_{j}L_{cls}(c_j,c_j^*) \\
	+ & W_h L_{reg}(h,h^*),
\end{aligned}
\end{equation}
where $W_p$, $W_s$, $W_c$, and $W_h$ are constant weights used for balancing. The regressions $L_{reg}$ and $L_{cls}$ are the Mean Squared Error (MSE) and Binary Cross Entropy (BCE) functions, respectively. The $L_p$ loss function measures how well adjusted is the polynomial $p_j$ (Equation \ref{eq:pol}) to the annotated points. Consider the annotated $x$-coordinates $\mathbf{x}^*_j=[x^*_{1,j}, \ldots, x^*_{N,j}]^T$ and $\mathbf{x}_j=[x_{1,j}, \ldots, x_{N,j}]$ where
\begin{equation}
	x_{i,j} = 
	\begin{cases}
	    p_j(y^*_{i,j}), & \text{if } |p_j(y^*_{i,j}) - x^*_{i,j} | > \tau_{loss}\\
	    0, & \text{otherwise.}
	\end{cases}
	\label{eq:auxvec}
\end{equation}
where $\tau_{loss}$ is an empirically defined threshold that tries to reduce the focus of the loss on points that are already well aligned. Such effect appears because the lane markings comprise several points with different sampling differences (i.e., points closer to the camera are denser than points further away). Finally, $L_p$ is defined as

\begin{equation}
	L_p(\{\mathcal{P}_j\}, \{\mathcal{L}^*_j\}) = L_{reg}(\mathbf{x}_j, \mathbf{x}^*_j).
	\label{eq:lp}
\end{equation}

\section{Experimental Methodology}

\label{sec:experiments}

\methodname{} was evaluated on publicly available which are introduced in this section. Following, the section describes the implementation details, the metrics, and the experiments performed.

\subsection{Datasets}
Three datasets were used to evaluate \methodname: TuSimple~\cite{tusimple-benchmark}, LLAMAS~\cite{llamas2019} and ELAS~\cite{berriel2017imavis}. For quantitative results, the widely-used TuSimple~\cite{tusimple-benchmark} was employed. The dataset has a total of 6,408 annotated images with a resolution of 1280$\times$720 pixels, and it is originally split in 3,268 for training, 358 for validation, and 2,782 for testing. For qualitative results, two other datasets were used: LLAMAS~\cite{llamas2019} and ELAS~\cite{berriel2017imavis}. The first is a large dataset, split into 58,269 images for training, 20,844 for validation, and 20,929 for test, with a resolution of 1280$\times$717 pixels. Both TuSimple and LLAMAS are datasets from the USA. Since neither the benchmark nor the test set annotations for LLAMAS are available yet, only qualitative results are presented. ELAS is a dataset with 16,993 images from various cities in Brazil, with a resolution of 640$\times$480 pixels. Since the dataset was originally proposed for a non-learning based method, it does not provide training/testing splits. Thus, we created those splits by separating 11,036 images for training and 5,957 for testing. The main difference between ELAS and the other two datasets is that in ELAS only the ego-lane is annotated. Nonetheless, the dataset also provides other types of useful information for the lane detection task, such as lane types (e.g., solid or dashed, white or yellow), but they are not used in this paper.

\subsection{Implementation details}
The hyperparameters for every experiment in this work were the same, except for the ablation study, where in each training one parameter was modified. 
For the backbone network, the EfficientNet-b0~\cite{tan2019efficientnet} was used. For the TuSimple training, data augmentation was applied with a probability of $\frac{10}{11}$. The transformations used were: rotation with an angle in degrees $\theta\sim\mathcal{U}(-10, 10)$, horizontal flip with a probability of 0.5, and a random crop with size 1152$\times$648 pixels. After the data augmentation, the following transformations were applied: a resize to 640$\times$360 pixels and then a normalization with ImageNet's~\cite{imagenet_cvpr09} mean and standard deviation. The Adam optimizer was used, along with the Cosine Annealing learning rate scheduler with an initial learning rate of 3e-4 and a period of 770 epochs. The training session ran for 2695 epochs, taking approximately 35 hours on a Titan V, with a batch size of 16 images, from a model pretrained on ImageNet~\cite{imagenet_cvpr09}. A third-order polynomial degree was chosen to be the default. For the loss function, the parameters $W_s=W_c=W_h=1$ and $W_p=300$ were used. The threshold $\tau_{loss}$ (Equation \ref{eq:auxvec}) was set to 20 pixels. In the testing phase, lane markings predicted with a confidence score $c_j < 0.5$ were ignored. For more details, the source code and trained models are publicly available\footnote{\url{https://github.com/lucastabelini/PolyLaneNet}}.

\subsection{Evaluation Metrics}
\label{sec:evalmetrics}
The metrics used to measure the proposed method's performance come from TuSimple's benchmark~\cite{tusimple-benchmark}. The three metrics are: accuracy $(Acc)$, false positive $(FP)$ and false negative $(FN)$ rates. For a predicted lane marking to be considered a true positive (i.e., a correct one), its accuracy, defined as
\begin{equation}
Acc(\mathcal{P}_j, \mathcal{L}^*_j) = \frac{1}{|\mathcal{L}^*_j|}\sum_{(x^*_{i,j}, y^*_{i,j}) \in \mathcal{L}^*_j}\textbf{1}[|p_j(y^*_{i,j}) - x^*_{i,j}| < \tau_{acc}]
\end{equation}
has to be equal to or greater than $\epsilon$. The values used for $\tau_{acc}$ and $\epsilon$ were 20 pixels and 0.85, respectively, the same ones used in TuSimple's benchmark. All the three reported metrics $(Acc,$ $FP$ and $FN)$ are reported as the average across all images of the average of each image.

Although TuSimple's metric has been widely used in the literature, it is too permissive w.r.t. local errors. To avoid relying on only such metric, we also used a metric proposed in~\cite{satzoda2014icpr}, which discusses several evaluation metrics of interest to the lane estimation process. The Lane Position Deviation (LPD) was proposed to better capture the accuracy of the model on both the near and far depths of view of the ego-vehicle. It is the error between the prediction and the ground-truth for the ego-lane. To define what are the ego-lanes (given that the dataset labels and our model are agnostic to this definition), we use a simple definition: the lane markings that are closer to the middle of the bottom part of the image are the ones that compose the ego-lane, i.e., one lane marking to the left and another one to the right.

In addition to metrics w.r.t. the quality of the predictions, we also report two speed-related metrics: frames-per-second (FPS) and MACs\footnote{For reference, roughly speaking, one MAC (Multiply-Accumulate) is equivalent to 2 FLOPS. MACs were computed using the following library: \url{https://github.com/mit-han-lab/torchprofile}.}. The frames-per-second provide a concrete assessment of how fast an implementation can run on a modern computer with a recent GPU, whereas MACs provide a more reliable way to compare different methods that might be running on different frameworks and setups. As discussed in~\cite{satzoda2014icpr}, analyzing the trade-off between computation efficiency and accuracy is also important. In this paper, we provide such an analysis by reporting the MACs of PolyLaneNet variants with different computational requirements in an ablation study.

\subsection{Quantitative Evaluation}

\textbf{State-of-the-art Comparison.}
The main quantitative experiment for the proposed method is the comparison against state-of-the-art methods using the same evaluation conditions. For that, the proposed method was used to train a model using a union of TuSimple's training and validation sets and then evaluated in its testing set. Four state-of-the-art methods were compared: SCNN~\cite{scnn2018aaai}, Line-CNN~\cite{linecnn2019tits}, ENet-SAD~\cite{enetsad2019iccv}, and FastDraw~\cite{fastdraw2019cvpr}. Besides prediction quality metrics, model speed w.r.t. FPS is also presented. For our model, we also reported the MACs.

\textbf{Polynomial Degree.}
In most lane marking detection datasets, it is clear that lane markings with a more accentuated curvature are rarer, while straight ones represent the majority of the cases. With this in mind, one might enquire: what would be the impact of modeling lane markings with polynomials of lower orders? To help answer this question, our method was evaluated using first- and second-order polynomials, instead of the default of third-order polynomials. Furthermore, we also show the permissiveness of the standard TuSimple's metric used by the literature by computing upper bounds for polynomials of different orders.

\textbf{Ablation Study.}
To investigate the impact of some of the decisions made for the proposed method, an ablation study was carried out, using only TuSimple's training set for training and the validation set for testing. For the model backbone $f(\cdot,\theta)$, ResNet~\cite{he2016deep} was evaluated, on two of its variants: ResNet-34 and 50. Another variant of EfficientNet was also evaluated, the EfficientNet-b1. Moreover, when training CNNs, in addition to the impact of the backbone, there is a trade-off when using different image input sizes. For example, if a smaller input size is used, the network forward will be faster, but information may be lost. To measure this trade-off~\cite{satzoda2014icpr} in the proposed method, two additional models were trained, one using an input size of $480\times270$ pixels, and the other using an input size of $320\times180$ pixels. Additionally, three other practical decisions were evaluated: (i) the impact of not sharing $h$ (i.e., having the end of each lane predicted individually), (ii) the use of a pre-trained model, by training from scratch instead of a model pretrained on ImageNet; and (iii) the impact of using data augmentation, by removing the online data augmentation, which reduces the variability seen by the model at training time.

\subsection{Qualitative Evaluation}
For qualitative results, an extensive evaluation was carried out. Using the model trained on TuSimple as a pretraining, three models were trained: two on ELAS, one with and one without lane marking type classification, and another on LLAMAS. On ELAS, the model was trained for 385 additional epochs (half of a period of the chosen learning rate scheduler, where the learning rate will be at a minimum). On LLAMAS, the model was trained for 75 additional epochs, an approximation to the number of iterations used on ELAS, as the training set of LLAMAS is around five times larger than the one of ELAS. The experiment with lane marking type classification is a straightforward extension of PolyLaneNet, in which a category is predicted for each lane showcasing how trivial it is to extend our model.

\section{Results}

First, we present the results of the comparison with the state-of-the-art. Then, the results of the ablation study are detailed and discussed. Finally, qualitative results are shown.

\begin{figure}
	\centering
	\subfigure{\includegraphics[width=0.49\linewidth]{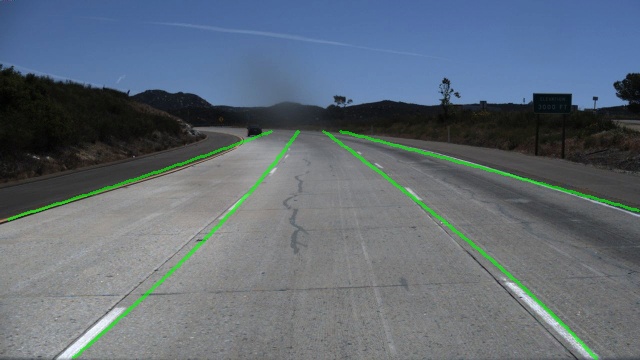}}
	\subfigure{\includegraphics[width=0.49\linewidth]{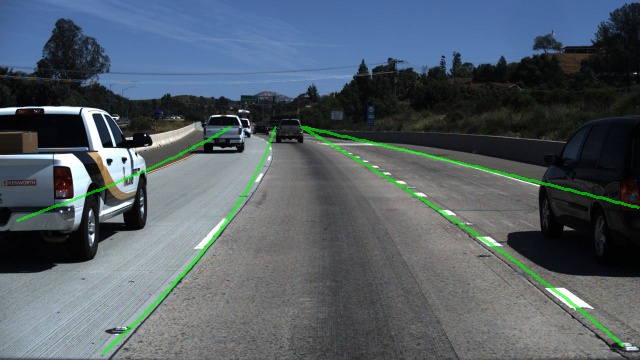}}
	\subfigure{\includegraphics[width=0.49\linewidth]{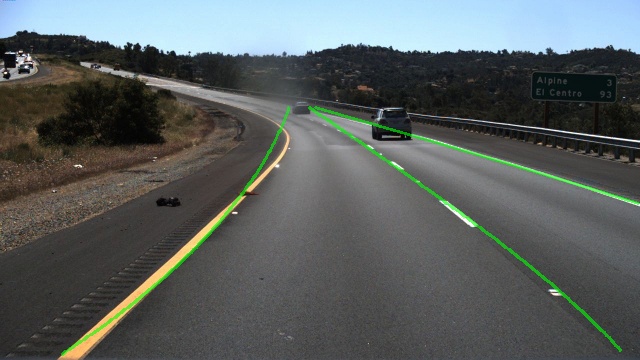}}
	\subfigure{\includegraphics[width=0.49\linewidth]{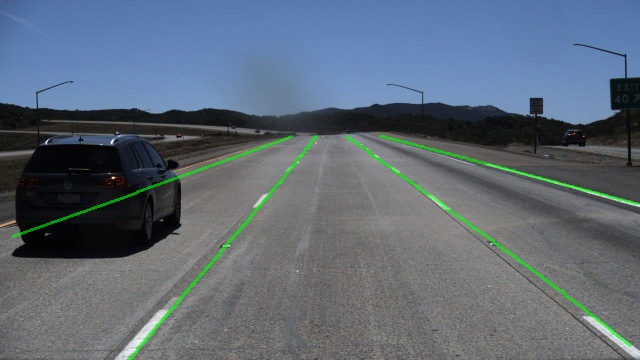}}
	\caption{Qualitative results of \methodname{} on TuSimple.}
	\label{fig:tusimple_qualitative}
\end{figure}

\textbf{State-of-the-art Comparison.}
The state-of-the-art results on the TuSimple dataset are presented in~\Cref{tab:sota}. As evidenced, \methodname{} results are competitive. Since none of the compared methods provide source codes that replicate their respective published results, it is very difficult to investigate situations where the other methods succeed that ours fail. On \Cref{fig:tusimple_qualitative}, some qualitative results of \methodname{} on TuSimple are shown. It is noticeable that \methodname{}'s predictions on parts of the lane marking closer to the camera (where more details can be seen) are very accurate. Nonetheless, on parts of the lane marking closer to the horizon, the predictions are less accurate. We conjecture that this might be a result of a local minimum, caused by the dataset's imbalance. Since most lane markings in the dataset can be represented fairly well with 1st order polynomials (i.e., lines), the neural network has a bias towards predicting lines, thus the poor performance on lane markings with accentuated curvature. 

\begin{table}[h]
	\centering
	\caption{State-of-the-art results on TuSimple.\protect\\ PP = Requires Post-Processing.}
	\label{tab:sota}
	\begin{tabular}{@{}lccc|cc|c@{}}
		\toprule
		\textbf{Method}                   & \textbf{Acc}      & \textbf{FP}       & \textbf{FN}        & \textbf{FPS} & \textbf{MACs} & \textbf{PP} \\ \midrule
		Line-CNN~\cite{linecnn2019tits}   & \textbf{96.87}\%  & \textbf{0.0442}   & 0.0197             & 30           &                &             \\
		ENet-SAD~\cite{enetsad2019iccv}   & 96.64\%           & 0.0602            & 0.0205             & 75           &                & \checkmark  \\
		SCNN~\cite{scnn2018aaai}          & 96.53\%           & 0.0617            & \textbf{0.0180}    & 7            &                & \checkmark  \\
		FastDraw~\cite{fastdraw2019cvpr}  & 95.20\%           & 0.0760            & 0.0450             & 90           &                & \checkmark  \\ \midrule
		PolyLaneNet                       & 93.36\%           & 0.0942            & 0.0933             & 115          & 1.748G          &             \\ \bottomrule
	\end{tabular}
\end{table}

\textbf{Polynomial Degree.}
In terms of the polynomial degree used to represent the lane marking, the small difference in accuracy when using lower-order polynomials shows how unbalanced the dataset is. Using 1st order polynomials (i.e., lines) decreased the accuracy by only 0.35 p.p. Although the dataset's imbalance certainly has an impact on this, another important factor is the metric used by the benchmark to evaluate a model's performance. The LPD metric~\cite{satzoda2014icpr}, however, is able to better capture the difference between the models trained using 1st order polynomials and the others. This can be further seen in \Cref{tab:upperbound}, which shows the maximum performance (i.e., the upper bound) of methods that represent lane markings as polynomials, measured by fitting polynomials on the test data itself. As it can be seen, TuSimple's metric does not punish predictions that are accurate only in parts of the lane marking closer to the car, wherein the image it will look almost straight (i.e., can be represented well by 1st order polynomials), since the thresholds may hide those mistakes. Meanwhile, the LDP metric clearly distinguishes the upper bounds, showing a clear difference even between the 4th and 5th degrees, in which  TuSimple's metrics are almost identical.

\begin{table}[h]
	\centering
	\caption{Ablation Study results on TuSimple Validation Set\protect\\ w.r.t. Polynomial Degree}
	\label{tab:ablation-study-poly-degree}
	\begin{tabular}{@{}ccccc|c@{}}
		\toprule
		\multicolumn{2}{c}{\textbf{Modification}}  & \textbf{Acc}  & \textbf{FP} & \textbf{FN} & \textbf{LPD}         \\ \midrule
		\multirow{3}{*}{Polynomial Degree} & 1st   & 88.63\%       & 0.2231      & 0.1865      & 2.532 \\
		                                   & 2nd   & \textbf{88.89\%}       & \textbf{0.2223}      & 0.1890      & 2.316 \\
		                                   & 3rd   & 88.62\%       & 0.2237      & \textbf{0.1844}      & \textbf{2.314} \\ \bottomrule
	\end{tabular}
\end{table}

\begin{table}[h]
\centering
\caption{TuSimple Performance Upperbound of Polynomials}
\label{tab:upperbound}
\begin{tabular}{@{}cccc|c@{}}
\toprule
\textbf{Polynomial Degree} & \textbf{Acc} & \textbf{FP} & \textbf{FN} & \textbf{LPD} \\ \midrule
1st                        & 96.22\%                          & 0.0393                          & 0.0367  & 1.512                        \\
2nd                        & 97.25\%                          & 0.0191                          & 0.0175  & 1.116                        \\
3rd                        & 97.84\%                          & 0.0016                          & 0.0014  & 0.732                        \\
4th                        & 98.00\%                          & 0.0000                          & 0.0000  & 0.497                        \\
5th                        & 98.03\%                          & 0.0000                          & 0.0000  & 0.382                        \\ \bottomrule
\end{tabular}
\end{table}

\textbf{Ablation Study.}
The ablation study results are shown in~\Cref{tab:ablation-study-backbone-inputsize}. EfficientNet-b1 achieved the highest accuracy, followed by EfficientNet-b0 and ResNet-34. Those results suggest that larger networks, such as ResNet-50, may overfit the data. Although EfficientNet-b1 achieved the highest accuracy, we chose not to use it in other experiments, as the accuracy gains are not significant nor consistent in our experiments. In addition, it is more computationally expensive (i.e., lower FPS, higher MACs, and longer training times). In regards to the input size, reducing it also means reducing the accuracy, as expected. In some cases, this accuracy loss may not be significant, but the speed gains may be. For example, using an input size of 480$\times$270 decreased the accuracy by only 0.55 p.p., but the model MACs decreased by 1.82 times.

\begin{table}[h]
	\centering
	\caption{Ablation Study results on TuSimple Validation Set\protect\\ w.r.t Backbone and Input Size}
	\label{tab:ablation-study-backbone-inputsize}
	\begin{tabular}{@{}ccccc|c@{}}
		\toprule
		\multicolumn{2}{c}{\textbf{Modification}}      & \textbf{Acc}      & \textbf{FP}     & \textbf{FN}     & \textbf{MACs (G)} \\ \midrule
		\multirow{4}{*}{Backbone}    & ResNet-34       & 88.07\%           & 0.2267          & 0.1953          & 17.154            \\
		                             & ResNet-50       & 83.37\%           & 0.3472          & 0.3122          & 19.135            \\
		                             & EfficientNet-b1 & \textbf{89.20\%}  & \textbf{0.2170} & \textbf{0.1785} & 2.583            \\
		                             & EfficientNet-b0 & 88.62\%           & 0.2237          & 0.1844          & 1.748            \\ \midrule
		\multirow{3}{*}{Input Size}  & 320$\times$180  & 85,45\%           & 0.2924          & 0.2446          & 0.396            \\
		                             & 480$\times$270  & 88.39\%           & 0.2398          & 0.1960          & 0.961            \\
		                             & 640$\times$360  & \textbf{88.62\%}  & \textbf{0.2237} & \textbf{0.1844} & 1.748            \\ \bottomrule
	\end{tabular}
\end{table}

As to the other ablation studies we carried out, one can see that sharing the top-y ($h$) is slightly better than not sharing. Moreover, training from a model pretrained on ImageNet seems to have a significant impact on the final result, as shown by the difference of 4.26 p.p. The same happens with data augmentation, as the model trained with more data has a significantly higher accuracy.

\begin{table}[h]
	\centering
	\caption{Ablation Study results on TuSimple Validation Set}
	\label{tab:ablation-study-others}
	\begin{tabular}{@{}ccllll@{}}
		\toprule
		\multicolumn{2}{c}{\textbf{Modification}} & \multicolumn{1}{c}{\textbf{Acc}} & \multicolumn{1}{c}{\textbf{FP}} & \multicolumn{1}{c}{\textbf{FN}}\\ \midrule
		\multirow{2}{*}{Top-Y Sharing}     & No              & 88,43\%  & 0.2126 & 0.1783 \\
		                                   & Yes             & 88.62\%  & 0.2237 & 0.1844 \\ \midrule
		\multirow{2}{*}{Pretraining}       & None            & 84,37\%  & 0.3317 & 0.2826 \\
		                                   & ImageNet        & 88.62\%  & 0.2237 & 0.1844 \\ \midrule
		\multirow{3}{*}{Data Augmentation} & None            & 78.63\%  & 0.4188 & 0.4048 \\
		                                   & 10$\times$      & 88.62\%  & 0.2237 & 0.1844 \\ \bottomrule
	\end{tabular}
\end{table}

\begin{figure*}[t]
	\centering
	\subfigure{\includegraphics[width=0.24\linewidth]{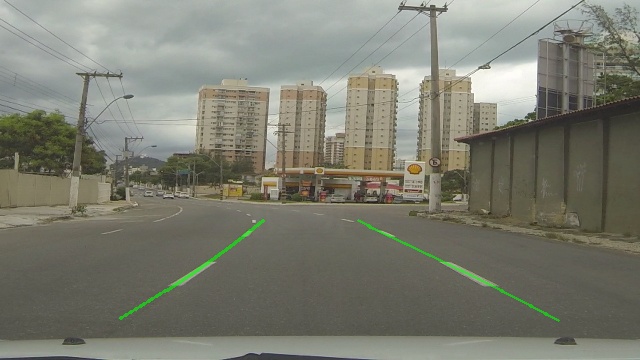}}
	\subfigure{\includegraphics[width=0.24\linewidth]{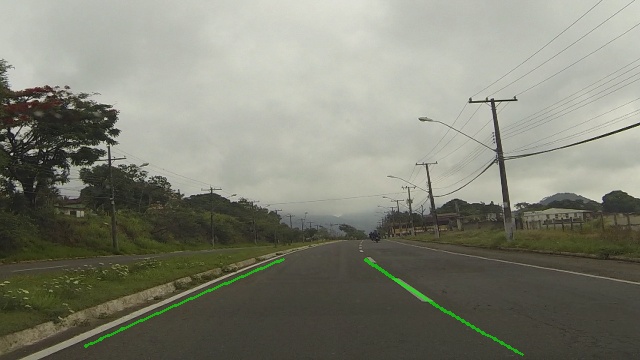}}
	\subfigure{\includegraphics[width=0.24\linewidth]{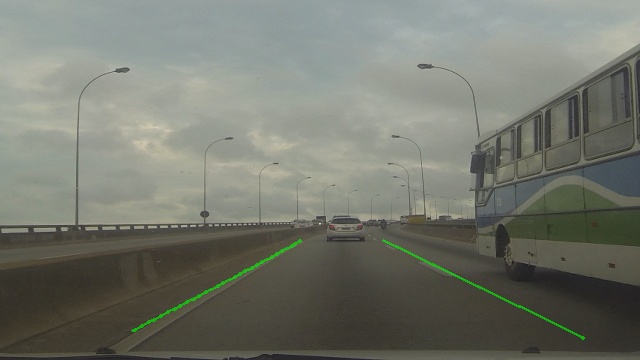}}
	\subfigure{\includegraphics[width=0.24\linewidth]{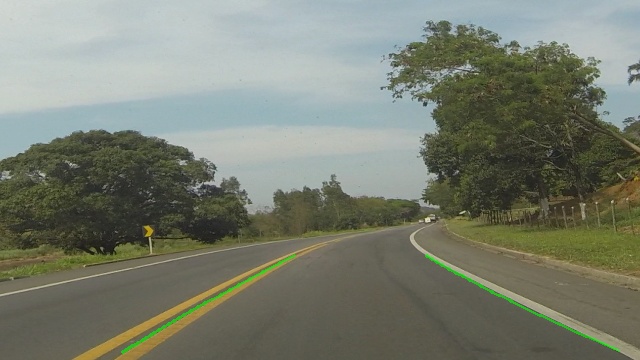}}
	\subfigure{\includegraphics[width=0.24\linewidth]{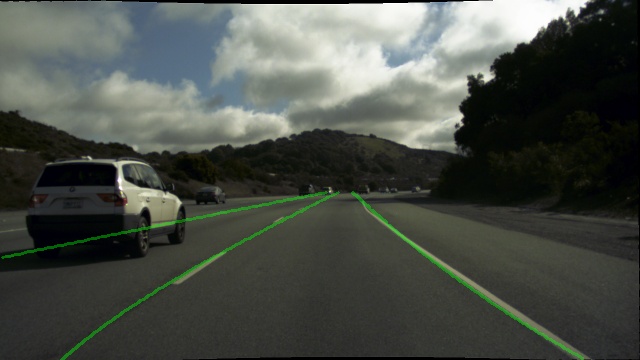}}
	\subfigure{\includegraphics[width=0.24\linewidth]{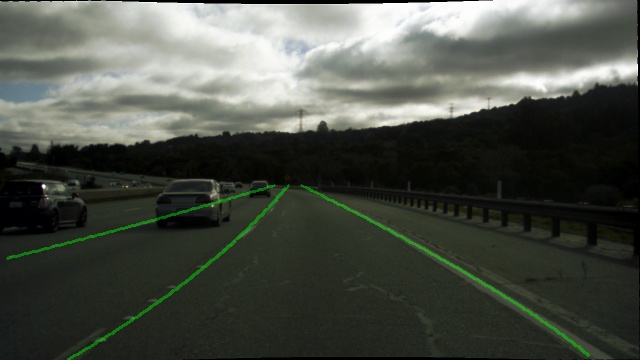}}
	\subfigure{\includegraphics[width=0.24\linewidth]{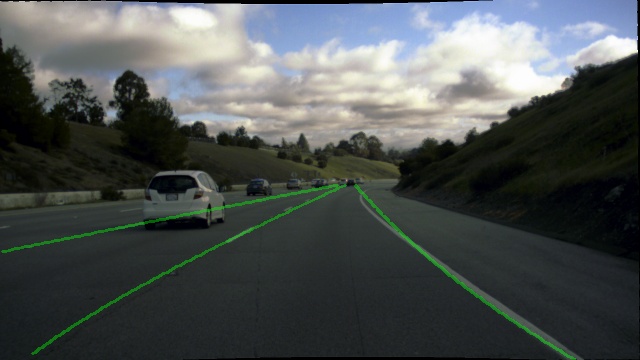}}
	\subfigure{\includegraphics[width=0.24\linewidth]{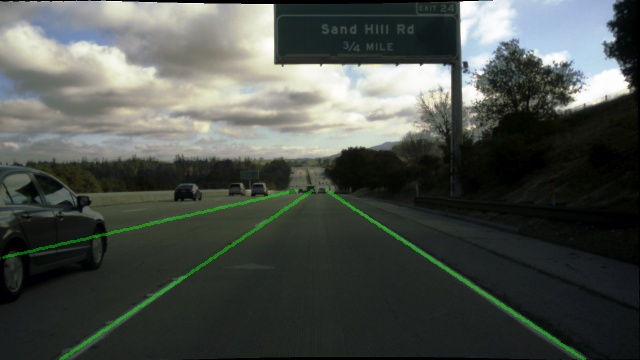}}
	
	\caption{Qualitative results of \methodname{} on ELAS (top row) and LLAMAS (bottom row).}
	\label{fig:qualitative-elas-llamas}
\end{figure*}

\textbf{Qualitative Evaluation.} A sample of the qualitative results on ELAS and LLAMAS are shown in~\Cref{fig:qualitative-elas-llamas}. For more extensive results, videos are available\footnote{Qualitative results (videos) on ELAS/LLAMAS: \url{https://www.youtube.com/playlist?list=PLm8amuguiXiJ2zKvcapUJI_ybyOFi9yz9}}. The results show that transfer learning works well on \methodname{}, since a smaller number of epochs was enough to obtain reasonable results in different datasets. However, in ELAS, there are many lane changes. In those situations, the model's accuracy decreased significantly. Since the images of those situations have a very different structure (e.g., the car is not heading towards the road direction), the low amount of samples in this situation may have not been enough for the model to learn it.

\section{Conclusion}
In this work, a novel method for lane detection based on deep polynomial regression was proposed. The proposed method is simple and efficient while maintaining competitive accuracy when compared to state-of-the-art methods. Although works with state-of-the-art methods with slightly higher accuracy exist, most do not provide source code to replicate their results, therefore deeper investigations on differences between methods are difficult. Our method, besides being computationally efficient, will be publicly available so that future works on lane markings detection have a baseline to start work and for comparison. Furthermore, we've shown problems on the metrics used to evaluate lane markings detection methods. For future works, metrics that can be used across different approaches to lane detection (e.g., segmentation) and that better highlights flaws in lane detection methods can be explored.
\section*{Acknowledgment}
This study was financed in part by the Coordenacão de Aperfeiçoamento de Pessoal de Nível Superior - Brasil (CAPES) - Finance Code 001, Conselho Nacional de Desenvolvimento Científico e Tecnológico (CNPq, Brazil), PIIC UFES and Fundação de Amparo à Pesquisa do Espírito Santo - Brasil (FAPES) – grant 84412844. We thank NVIDIA for providing GPUs used in this research.

\balance

\bibliographystyle{IEEEtran}
\bibliography{IEEEabrv,refs}

\end{document}